\relax
\documentclass[letterpaper]{article} 
\usepackage{aaai22}  
\usepackage{times}  
\usepackage{helvet}  
\usepackage{courier}  
\usepackage[hyphens]{url}  
\usepackage{graphicx} 
\usepackage{natbib}  
\usepackage{caption} 
\DeclareCaptionStyle{ruled}{labelfont=normalfont,labelsep=colon,strut=off} 
\usepackage{algorithm}
\usepackage{algorithmic}
\usepackage{xcolor}         
\usepackage{amsmath}

\usepackage{graphicx}
\usepackage{amssymb}
\usepackage{pifont}
\newcommand{\cmark}{\ding{51}}%
\newcommand{\xmark}{\ding{55}}%

\usepackage{booktabs}
\usepackage{stfloats}
 
\usepackage{multirow}

\usepackage{newfloat}
\usepackage{listings}
\floatstyle{ruled}
\newfloat{listing}{tb}{lst}{}
\floatname{listing}{Listing}

\title{BioGrad: Biologically Plausible Gradient-Based Learning for Spiking Neural Networks}
\author {
    Guangzhi Tang,
    Neelesh Kumar,
    Ioannis Polykretis,
    and Konstantinos P. Michmizos
}
\affiliations {
    Department of Computer Science, Rutgers University\\
    \{gt235, nk525, ip211, michmizos\}@cs.rutgers.edu
}

\usepackage{bibentry}

\begin{document}

\maketitle

\begin{abstract}
Spiking neural networks (SNN) have started to deliver energy-efficient, massively parallel, and low-latency solutions to AI problems, facilitated by the emerging neuromorphic hardware. To harness these computational benefits, SNN need to be trained by learning algorithms that adhere to brain-inspired neuromorphic principles, namely event-based, local, and online computations. However, the state-of-the-art SNN training algorithms are based on backpropagation that does not follow the above neuromorphic computational principles. Due to its limited biological plausibility, the application of backprop to SNN requires non-local feedback pathways for transmitting continuous-valued errors, and relies on gradients from future timesteps. The recent introduction of biologically plausible modifications to backprop has helped overcome several of its limitations, but limits the degree to which backprop is approximated, which hinders its performance. Here, we propose a biologically plausible gradient-based learning algorithm for SNN that is functionally equivalent to backprop, while adhering to all three neuromorphic computational principles. We introduced multi-compartment spiking neurons with local eligibility traces to compute the gradients required for learning, and a periodic "sleep" phase to further improve the approximation to backprop during which a local Hebbian rule aligns the feedback and feedforward weights. Our method achieved the same level of performance as backprop with multi-layer fully connected SNN on MNIST ($98.13\%$) and the event-based N-MNIST ($97.59\%$) datasets. We then deployed our learning algorithm on Intel's Loihi neuromorphic processor to train a 1-hidden-layer network for MNIST, and obtained $93.32\%$ test accuracy while consuming $400$ times less energy per training sample than BioGrad on GPU. Our work demonstrates that optimal learning is feasible in neuromorphic computing, and further pursuing its biological plausibility can better capture the computational benefits of this emerging computing paradigm.
\end{abstract}

Neuromorphic computing leverages the computational benefits of the brain by capturing brain-inspired computational principles. Such computational principles are seamlessly expressed in spiking neural networks (SNN) that run on the emerging neuromorphic processors \citep{davies2018loihi,merolla2014million,schemmel2010wafer,furber2014spinnaker, akida2019brain}. Mimicking their biological counterparts, neurons in SNN communicate through discrete localized events, called spikes, which give rise to the three main neuromorphic principles. First, \textit{spike-based computations} offer energy-efficient solutions to a wide variety of tasks \citep{davies2021advancing, esser2016convolutional,tang2019spiking}. Second, \textit{local information processing} enables asynchronous computations, which allows for massive parallelism \citep{davies2018loihi,davies2021advancing}. Third, neurons perform \textit{rapid online computations} without requiring information from future timesteps, resulting in low-latency solutions \citep{davies2021advancing,stagsted2020towards,taunyazov20event}. Adhering to these computational principles is of paramount importance for training SNNs that harness the computational benefits, as has been demonstrated by their integration with the widely used Hebbian learning rules \citep{davies2018loihi,imam2020rapid}. However, despite their biological plausibility, correlation-based learning rules have limited expressiveness, and cannot approximate complex functions well \citep{legenstein2005can}. This lack of SNN learning rules that are both effective and incorporate the above principles, hinders the neuromorphic technology from going mainstream. 

Backpropagation-based gradient descent has been remarkably effective in training deep neural networks, achieving near-human performance levels on several challenging tasks \citep{lecun2015deep}. Propelled by this success, several variants of backpropagation algorithms have recently been proposed to directly train SNN by backpropagating both the spatial gradients from downstream layers and the temporal gradients from future timesteps \citep{wu2018spatio, shrestha2018slayer, belleclong, tang2020deep}. While backprop achieves state-of-the-art performance in SNN \citep{wu2018spatio}, its limited biological plausibility forces it to violate the above computational principles needed for its direct implementations on neuromorphic chips. First, backprop requires separate feedback pathways to transmit the error signals; The feedback pathways need knowledge of non-local downstream synaptic connections to compute synaptic weight updates accurately- a problem known as weight-transport \citep{grossberg1987competitive, lillicrap2016random}. Second, since the weight updates require gradients to be backpropagated from future timesteps, learning cannot be directly implemented online in a neuromorphic hardware. Third, the algorithm propagates continuous-valued errors, which does not align with the chips' mode of discrete event computations. Therefore, there is a need for a gradient-based learning rule for SNN that a) is as effective as backprop, while b) following the computational principles so that it can be directly implemented on neuromorphic hardware.

Interestingly, recent efforts have sought to tackle some of the above limitations of backpropagation separately, by proposing biologically plausible modifications to it. In particular, \citep{lillicrap2016random} addressed the need for non-local information due to weight-transport by showing that random feedback weights in a network can support error-backpropagation. Building upon this and inspired by the morphology of the neocortical pyramidal neurons, \citep{guerguiev2017towards,neftci2017event} proposed networks comprising of multi-compartment neurons that receive the feedforward and feedback information in segregated compartments. This partially addressed the need for different feedforward and feedback pathways. Likewise, \citep{bellec2020solution} addressed the dependence on information from future timesteps by using eligibility traces that reflect intermediate neuronal states. However, the introduction of biologically plausible modifications to backprop in the above methods limits the degree to which backprop can be approximated, resulting in decreased performance. These studies suggest that further endowing backpropagation with biological plausibility will allow it to adhere to the above neuromorphic computational principles, resulting in a learning rule for SNNs that is both efficient and generalizable.

In this work, we propose the \textbf{Bio}logically plausible \textbf{Grad}ient-based learning (\textbf{BioGrad}) for SNN that is functionally equivalent to backprop and follows the neuromorphic computational principles. Our learning algorithm utilized a multi-compartment neuron model, having distinct somatic and apical compartments, that allowed for simultaneous integration of feedforward and feedback information in the same neuron. Our learning rule closely approximated the backprop gradients using local eligibility traces, and was further improved by a periodic "sleep" phase, wherein the feedback weights were updated to match the feedforward weights using random inputs and a local Hebbian rule. We demonstrate our method's applicability to multilayer architectures in learning task-relevant higher-order representations, by achieving the same level of performance as backprop with fully-connected networks on two datasets, namely the MNIST ($98.13\%$) and the event-based N-MNIST ($97.59\%$). We then deployed our learning algorithm on Intel's Loihi neuromorphic processor to train a 1-hidden-layer network for MNIST, and obtained $93.32\%$ test accuracy while consuming $400$ times less energy per training sample than BioGrad on GPU. This provides further evidence that our method is applicable to  neuromorphic hardware.

\section{Related Works}
Several studies are now attempting to reconcile backpropagation with biological plausibility \citep{kolen1994backpropagation,lillicrap2016random,nokland2016direct}. For instance, \citep{lillicrap2016random,nokland2016direct} address the weight-transport problem by using random feedback weights to backpropagate the errors. This mechanism transmits useful teaching signals through a process known as feedback alignment, and achieves competitive performance on a variety of tasks when compared to backprop. However, these methods require a separate feedback pathway to transmit the error signals, which makes the weight updates non-local.

The need for separate feedback pathways is alleviated in \citep{zhao2020glsnn} by not only propagating errors directly from the output layer to hidden layers~\citep{nokland2016direct}, but also employing multi-compartment neurons that receive feedforward and feedback information in segregated compartments. However, these methods propagate continuous-valued errors, violating the event-based nature of computations. Networks with spike-based error computations are proposed in \citep{guerguiev2017towards,neftci2017event,shrestha2019approximating}, resulting in fully event-based training. But these methods require separate phases or more timesteps for error computation due to the use of static feedback weights, eliminating the advantage of low latency for neuromorphic computations. Our method circumvented this limitation using online event-based error computations and requiring fewer timesteps as a result of closely approximating the backprop gradients.

The biologically plausible modifications such as event-based error computations and feedback alignment limit the degree to which backprop is approximated, resulting in decreased performance. Improvements to feedback alignment are suggested in \citep{bellec2020solution, akrout2019deep,guerguiev2019spike} where local learning rules update the feedback weights using either the weight-mirroring principle or causal inference, so that they better align with the feedforward weights. However, the methods in \citep{akrout2019deep, guerguiev2019spike} are proposed for training non-spiking networks. Moreover, \citep{akrout2019deep} requires layer-wise propagation of errors which makes the method non-local. Likewise, \citep{bellec2020solution} relies on computation and propagation of continuous-valued errors, hence violating the event-based principle of neuromorphic computations. In addition, the adaptive version of their algorithm with feedback weight updates is based on \citep{kolen1994backpropagation} which requires a separate feedback pathway to transmit the error signals and violates the locality principle of neuromorphic computations. We overcame these limitations by introducing a sleep phase which aligned the feedback and the feedforward weights for a multi-layer SNN while following the neuromorphic principles.

\section{Biologically Plausible Gradient-Based Learning for SNN}
\begin{figure*}
    \centering
    \includegraphics[]{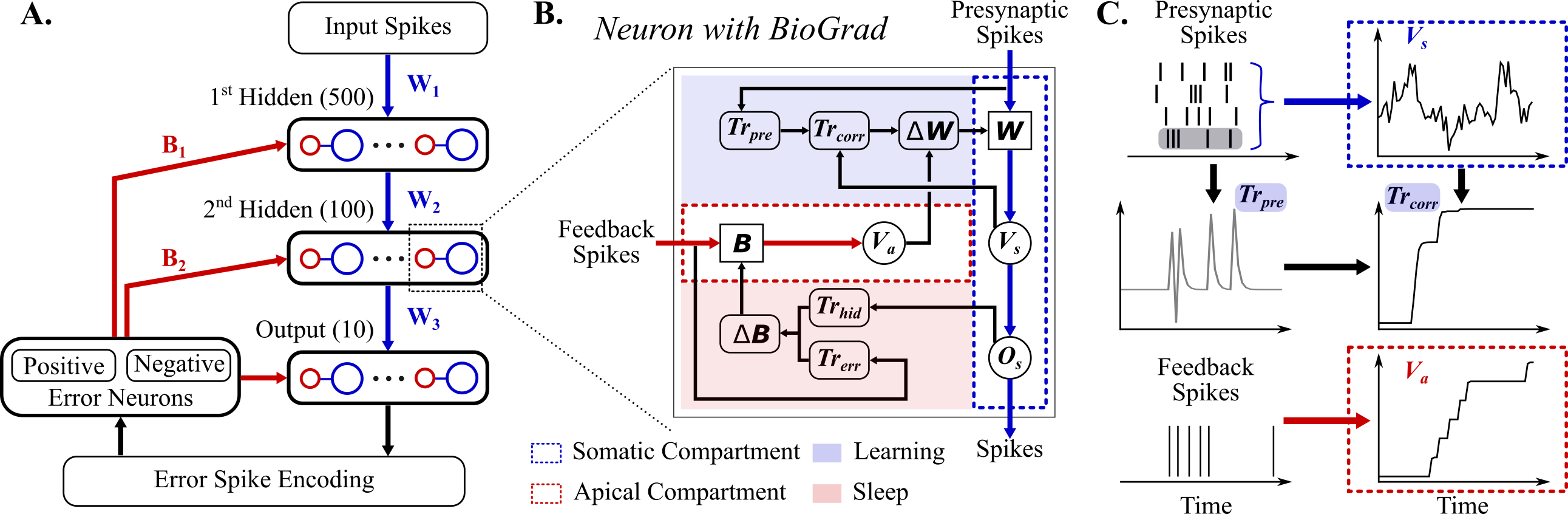}
    \caption{\textbf{A. }A fully connected 3-layer SNN consisting of multi-compartment neurons with apical (red circle) and somatic (blue circle) compartments. The somatic compartment integrated feedforward inputs (blue arrows), while the apical compartment integrated top-down feedback received directly from the error neurons (red arrows). \textbf{B. }Neuron with BioGrad that updated its feedforward and feedback weights using local rules based on neuronal states reflected in the eligibility traces. \textbf{C. } Illustration of how the presynaptic and feedback spikes determined the compartment voltages and eligibility traces.}
    \label{fig:fig1}
\end{figure*}
We propose a learning algorithm for SNNs (Fig. \ref{fig:fig1}) that is functionally equivalent to backpropagation and follows the main neuromorphic computational principles. Our proposed method utilized a multi-compartment neuron model consisting of a spiking somatic compartment and a non-spiking apical compartment, in accordance with the morphology of the neocortical pyramidal neurons~\citep{manita2015top}. This allowed for a separate but simultaneous integration of feedforward and feedback information within the same neuron. Specifically, the somatic compartment integrated the feedforward presynaptic inputs into its membrane voltage $\mathbf{v_s}$, and fired a spike whenever the membrane voltage exceeded a threshold. At the same time, the apical compartment received top-down feedback from the error neurons and integrated it into its membrane voltage $\mathbf{v_a}$. To update the feedforward weights, we used the states of these compartments to estimate the gradients of the loss with respect to the weights. Moreover, we introduced a periodic sleep phase where the feedback weights were updated in an unsupervised manner using a local Hebbian rule and random inputs. 

\subsection{Network Architecture}

We used the compartmental neurons described above to build a fully connected multi-layer SNN. During forward propagation, the presynaptic spikes to the $i^{th}$ layer at the $t^{th}$ timestep, $\mathbf{o_s}^{(i-1)(t)}$, were multiplied with the synaptic weights $\mathbf{W}^{(i)}$, and integrated into the membrane voltage of the somatic compartment $\mathbf{v_s}^{(i)(t)}$, following the dynamics of the leaky-integrate-and-fire neuron model. 
\begin{equation}
    \mathbf{v_s}^{(i)(t)} = d_v\cdot \mathbf{v_s}^{(i)(t-1)} + \mathbf{W}^{(i)}\mathbf{o_s}^{(i-1)(t)}
    \label{eq1}
\end{equation}
where $d_v$ is the voltage decay factor for the somatic compartment.  

When $\mathbf{v_s}^{(i)(t)}$ exceeded the threshold $V_{th}$, the somatic compartment generated a spike, which was then propagated to neurons in the next layer. Upon spiking, $\mathbf{v_s}^{(i)(t)}$ was reset to resting potential. At every timestep, we aggregated the spikes of the output layer neurons to compute an estimate of the classification error $\mathbf{e}^{(t)}$ as per equation (\ref{eq3}). 
\begin{equation}
\begin{gathered}
\mathbf{Output}^{(t)}= \mathbf{Output}^{(t-1)} + \mathbf{o_s}^{(out)(t)}\\
\mathbf{e}^{(t)} = Softmax(\mathbf{Output}^{(t)}) - label\_one\_hot
\end{gathered}
\label{eq3}
\end{equation}
where $label\_one\_hot$ is the one-hot-vector representation of the ground-truth label. At the end of the sample presentation time $T$, the output neuron with the highest number of spikes represented the predicted label. For an event-based computation of the error, we first encoded the errors $\mathbf{e}^{(t)}$ into spikes that drove the error neurons. Since spikes are unsigned, we used positive and negative error neurons to represent the differently signed errors. The error neuron spikes $\mathbf{o_e}^{(pos)(t)}, \mathbf{o_e}^{(neg)(t)}$ were then sent directly to the apical compartments  of each layer, and integrated into the apical voltage $\mathbf{v_a}$ after being multiplied with the feedback weights $\mathbf{B}^{(i)}$.
\begin{equation}
    \mathbf{v_a}^{(i)(t)} = \mathbf{v_a}^{(i)(t-1)} + \mathbf{B}^{(i)}\mathbf{o_e}^{(pos)(t)} - \mathbf{B}^{(i)}\mathbf{o_e}^{(neg)(t)}
    \label{eq4}
\end{equation}

The $i^{th}$ layer feedback weights $\mathbf{B}^{(i)}$  were initialized to products of feedforward weights as per equation (\ref{eq5}), and were updated during the sleep phase.
\begin{equation}
    \mathbf{B}^{(i)}_{init} = \prod_{j=i+1}^{K} \mathbf{W^{(j)}_{init}}^\top
    \label{eq5}
\end{equation}
where $K$ is the number of layers in the network and $\top$ denotes matrix transpose. 

\begin{figure*}
    \centering
     \includegraphics{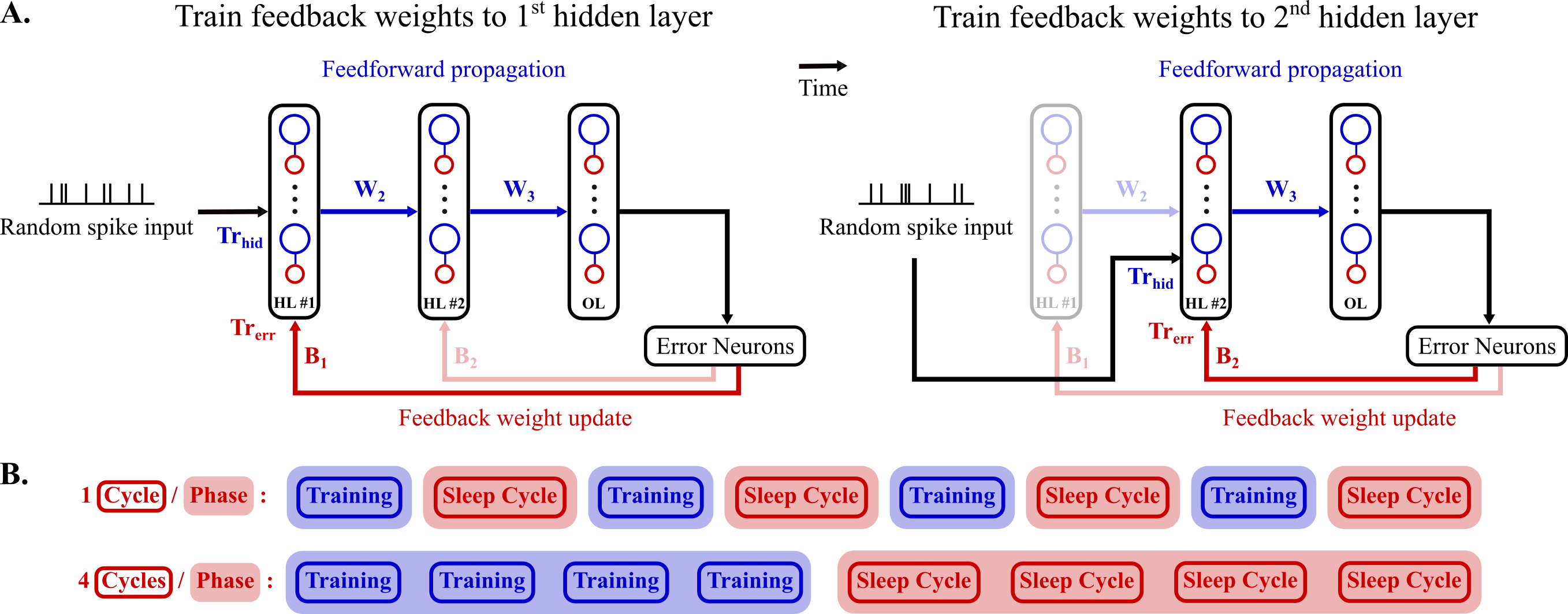}
    \caption{\textbf{A.} Illustration of a single sleep cycle. 
    Feedback weights to each hidden layer were updated sequentially using two traces: the response of the layer to a random input was recorded in $\mathbf{Tr_{hid}}$ and was propagated through the network; the network output was fed back to the apical compartments of the layer via error neurons and was recorded in $\mathbf{Tr_{err}}$. \textbf{B.} Example of sleep phase frequencies: sleep phases comprising of 4 (1) cycles occurring after every 4 (1) batches of training samples. }
    \label{fig:fig2}
\end{figure*}

\subsection{Synaptic Learning Rule}
We formulated a gradient-based online local learning rule to train the feedforward weights of the SNN. Our learning rule utilized two eligibility traces per synapse: a presynaptic spike trace $\mathbf{Tr_{pre}}$, and a correlation trace $\mathbf{Tr_{corr}}$. The trace $\mathbf{Tr_{pre}}$ reflected the history of the spiking activity of the presynaptic neuron. It is incremented with every presynaptic spike, and decays with a factor that depends on the state of the postsynaptic somatic compartment.
\begin{equation}
\begin{gathered}
\mathbf{Tr_{pre}}^{(i)(t)} = d_v\cdot \mathbf{Decay}^{(i)(t)} \cdot \mathbf{Tr_{pre}}^{(i)(t-1)} + \mathbf{o_s}^{(i-1)(t)}\\
\mathbf{Decay}^{(i)(t)} = 1-\mathbf{o_s}^{(i)(t-1)}-\mathbf{v_s}^{(i)(t-1)}\cdot z(\mathbf{v_s}^{(i)(t-1)})
\end{gathered}
\end{equation}
where $z$ is a rectangular function for computing the pseudo-gradient~\citep{wu2018spatio} of a spike.

The trace $\mathbf{Tr_{corr}}$ captures the correlation between the presynaptic and postsynaptic neuron activities. 
\begin{equation}
    \mathbf{Tr_{corr}}^{(i)(t)} = \mathbf{Tr_{corr}}^{(i)(t-1)} + \mathbf{Tr_{pre}}^{(i)(t)}\cdot z(\mathbf{v_s}^{(i)(t-1)})
\end{equation}

The computed traces were used to update the weights as follows:
\begin{equation}
    \Delta \mathbf{W}^{(i)} = -\eta\cdot \frac{\mathbf{v_a}^{(i)(T)}}{T-t_{error}}\cdot \mathbf{Tr_{corr}}^{(i)(T)}
\end{equation}
where $\eta$ is the learning rate and $t_{error}$ is the starting timestep at which the spike-encoded errors were fed to the error neurons. Our learning rule is mathematically equivalent to an approximated form of spatiotemporal backpropagation for SNN~\citep{wu2018spatio, shrestha2018slayer, belleclong}. In this approximation (equation (\ref{eq8})), the gradient of the loss $L$ with respect to the weights comprises of a spatial and a temporal gradient. The spatial gradient is computed by backpropagating the loss gradient at the output layer,  $\frac{\partial L}{\partial Output(T)}$, through feedback weights $\mathbf{B}$. The temporal gradient is computed by accumulating the local gradient of the neuron over all timesteps and thus requires information from the future timesteps.  
\begin{multline}
    \frac{\partial L}{\partial \mathbf{W}^{(i)}} \approx \mathbf{B}^{(i)}\frac{\partial L}{\partial Output(T)} .\\
\sum_{t=1}^T\frac{\partial \mathbf{v_s}^{(i)(t)}}{\partial \mathbf{W}^{(i)}}\sum_{x=t}^T \frac{\partial \mathbf{o_s}^{(i)(x)}}{\partial \mathbf{v_s}^{(i)(x)}} \prod_{y=1}^{x-1} d_v\cdot \mathbf{Decay}^{(i)(y)}
\label{eq8}
\end{multline}

To establish an equivalence with our learning rule, we note that the spatial gradient corresponds to the membrane voltage of the apical compartment $\mathbf{v_a}$, and the temporal gradient corresponds to the correlation trace $\mathbf{Tr_{corr}}$ (full mathematical proof in Appendix A).



\subsection{Sleep Phase for updating feedback weights}
While event-based feedback alignment ameliorates the weight transport problem, it approximates backprop only to a certain degree due to the use of static feedback weights \citep{neftci2017event}. To further improve this approximation under limited time steps, we introduced a sleep phase (Fig. \ref{fig:fig2}) in which the feedback weights for each hidden layer were updated sequentially in an unsupervised manner, in accordance with all the neuromorphic computational principles. The updates followed the weight mirroring principle~\citep{akrout2019deep} to allow the feedback weights $\mathbf{B}$ to grow in the direction of feedforward weights $\mathbf{W}$ (Appendix B). Specifically, to update the feedback weights $\mathbf{B^{(i)}}$ for a hidden layer $i$, we drove layer $i$ to generate independent and zero-mean outputs in the form of random positive and negative spikes $\mathbf{o_{pos}}$, $\mathbf{o_{neg}}$. Then, we propagated these spikes through the rest of the network: the activations of layer $i+1$ were computed as per equations (\ref{eq10}), while the activations of the subsequent layers were computed as per equations (\ref{eq1}). 
\begin{equation}
    \mathbf{v_s}^{(i+1)(t)} = d_v\cdot \mathbf{v_s}^{(i+1)(t-1)} + \mathbf{W}^{(i+1)}(\mathbf{o_{pos}}^{(t)} - \mathbf{o_{neg}}^{(t)})
    \label{eq10}
\end{equation}

The output layer spikes were then directly injected to the error neurons. During the propagation, we maintained two traces per neuron in layer $i$: $\mathbf{Tr_{hid}}^{(i)}$ to record the postsynaptic somatic activity of the hidden layer $i$, and $\mathbf{Tr_{err}}^{(i)}$ to record the activity of the error neurons received as a presynaptic input to the apical compartments of layer $i$.   
\begin{equation}
\begin{gathered}
\mathbf{Tr_{hid}}^{(i)(t)} = \mathbf{Tr_{hid}}^{(i)(t-1)} + (\mathbf{o_{pos}}^{(t)} - \mathbf{o_{neg}}^{(t)})\\
\mathbf{Tr_{err}}^{(i)(t)} = \mathbf{Tr_{err}}^{(i)(t-1)} + \mathbf{o_e}^{(t)}
\end{gathered}
\end{equation}
where $\mathbf{o_e}$ represents the spikes of either positive or negative error neurons. 

The feedback weights were then updated using a Hebbian-like local learning rule with a decay that resembled Oja's rule ~\citep{oja1982simplified} to prevent the uncontrollable growth of $\mathbf{B}$.  
\begin{equation}
\begin{aligned}
\Delta \mathbf{B}^{(i)} &= \beta\cdot \mathbf{Tr_{err}}^{(i)(T_{sleep})}\cdot (\mathbf{Tr_{hid}}^{(i)(T_{sleep})}  - \\
&\mathbf{Tr_{err}}^{(i)(T_{sleep})}\cdot  \mathbf{B}^{(i)})
\end{aligned}
\end{equation}

where $\beta$ is the sleep learning rate and $T_{sleep}$ is the presentation time for the random inputs. 

In principle, the network could be put to sleep at any time, at which point training was suspended. In practice, we introduced a periodic sleep phase after every fixed number of training samples (Fig. \ref{fig:fig2}B). Each sleep phase consisted of one or more sleep cycles. In each cycle, feedback weights for every layer were updated once depending upon the random inputs applied to the layers during that cycle.





\section{Experiments and Results}
We compared our method against backprop-based baselines to demonstrate their functional equivalence. We also performed ablation studies to show the effectiveness of the sleep phase in training the SNN. Lastly, we deployed our method on Intel's Loihi neuromorphic processor to demonstrate its on-chip learning capabilities.

\subsection{Experimental setup}
\begin{table*}[t]
  \caption{Comparing our method against other biologically plausible methods on MNIST}
  \label{mnist-brain-inspire}
  \centering
  \begin{tabular}{ccccccl}
    \toprule
    Method & Hidden layers & Spiking & Local & Online & Timesteps & Accuracy \\
    \midrule
    ~\citep{zhao2020glsnn} & 800-800-800 & \xmark & \xmark & \cmark & 10 & 98.62\% \\
    ~\citep{lillicrap2016random} & 1500-1500 & \xmark & \cmark & \cmark & 50 & 98.20\% \\
    ~\citep{tavanaei2019bp} & 500-150  & \cmark & \xmark & \cmark & 50 & 97.20\% \\
    ~\citep{shrestha2019approximating} & 500-500 & \cmark & \cmark & \xmark & 200 & 96.80\% \\
    ~\citep{guerguiev2017towards} & 500-100 & \cmark & \cmark & \xmark & 100 & 96.80\% \\
    ~\citep{neftci2017event} & 500-500-10 & \cmark & \cmark & \xmark & 250 & 97.98\% \\
    \textbf{BioGrad} & 500-100 & \cmark & \cmark & \cmark & 20 & \textbf{98.13\%}\ $\pm$\ \textbf{0.10\%} \\
    \bottomrule
  \end{tabular}
\end{table*}
We benchmarked our method on the MNIST \citep{lecun1998gradient} and the event-based N-MNIST \citep{orchard2015converting} datasets. We encoded each pixel in an MNIST image into a Poisson spike train whose firing rate depended upon the pixel intensity. The NMNIST dataset encodes each MNIST image in the form of spikes captured through the saccade movements of a DVS camera~\citep{orchard2015converting}. The resulting spatiotemporal dynamics pose a different challenge for the SNN than the static MNIST dataset. Both the MNIST and N-MNIST datasets consisted of 60,000 training (50,000 train + 10,000 validation) and 10,000 testing samples. We presented each MNIST sample for 20ms with a 1ms resolution, and each N-MNIST sample for 300 ms with a 5ms resolution. 

We trained fully connected SNN consisting of two hidden layers and one output layer (Fig. \ref{fig:fig1}A). As baselines, we implemented the state-of-the-art SNN backprop algorithm, STBP~\citep{wu2018spatio}, with stochastic gradient descent (SGD) and Adam optimizers. For fair comparisons, all the hyperparameters (Appendix C) for training the baselines except the learning rate were the same as the ones in our method. The learning rate was tuned separately for each method to maximize their performance. Further, to limit the effect of random initialization, we trained five networks initialized with five random seeds for all methods. Each network was trained for 100 epochs where 1 epoch corresponds to 1 pass over the entire training set. For evaluation, we picked the test accuracy corresponding to the highest validation accuracy for each random seed. We report the mean and standard deviation of the test accuracies computed over all the random seeds.

\subsection{Benchmarking BioGrad against backpropagation}
We compared the performance of our method against backprop on both datasets (Table \ref{main-result}). Our method fared better than backprop with SGD. However, it performed slightly worse than backprop with Adam on the more complex N-MNIST dataset, where the momentum and adaptive learning rate perhaps aiding the optimization. The slightly better performance of our method over backprop with SGD can be attributed to two possible reasons. First, the slight misalignment between feedforward and feedback weights introduced additional noise in the gradients which may have aided generalization~\citep{neelakantan2015adding, zhu2018anisotropic}. Second, the changes in the magnitudes of feedback weight due to sleep may have induced a similar effect as a learning rate scheduler, thereby improving performance~\citep{bengio2012practical}. Overall, the benchmarking results suggest that our method is functionally equivalent to backprop. 

\begin{table}[b]
\small

  \caption{Comparing BioGrad against backpropagation}
  \label{main-result}
  \centering
  \begin{tabular}{l c c}
    \toprule
    Method     & MNIST     & N-MNIST \\
    \midrule
    Backprop (SGD) & 97.49\%\ $\pm$\ 0.12\%  & 97.38\%\ $\pm$\ 0.07\%\\
    Backprop (Adam) & 98.10\%\ $\pm$\ 0.02\% & 98.31\%\ $\pm$\ 0.12\%      \\
    \textbf{BioGrad} & \textbf{98.13\%}\ $\pm$\ \textbf{0.10\%} & \textbf{97.59\%}\ $\pm$\ \textbf{0.12\%}  \\
    \bottomrule
  \end{tabular}
\end{table}

\subsection{Comparing BioGrad against other biologically plausible approaches}
Our method compared favorably against other biologically plausible gradient-based methods (Table \ref{mnist-brain-inspire}) with similar network architecture. Interestingly, incorporating the neuromorphic principles comes at the cost of a decrease in performance for the existing methods. For example, methods that employ continuous feedback computations~\citep{zhao2020glsnn,lillicrap2016random} outperformed those that have event-based feedback~\citep{guerguiev2017towards, tavanaei2019bp,shrestha2019approximating}. However, our method achieved a high level of performance without sacrificing any of the neuromorphic principles. This is because our learning rule closely approximated backprop by being mathematically equivalent to a modification of it (Appendix A). In addition, the sleep phase updated the feedback weights to better align with the feedforward weights, further improving the approximation. 

\subsection{Advantages of learning with hidden layers}
\begin{figure*}
    \centering
    \includegraphics{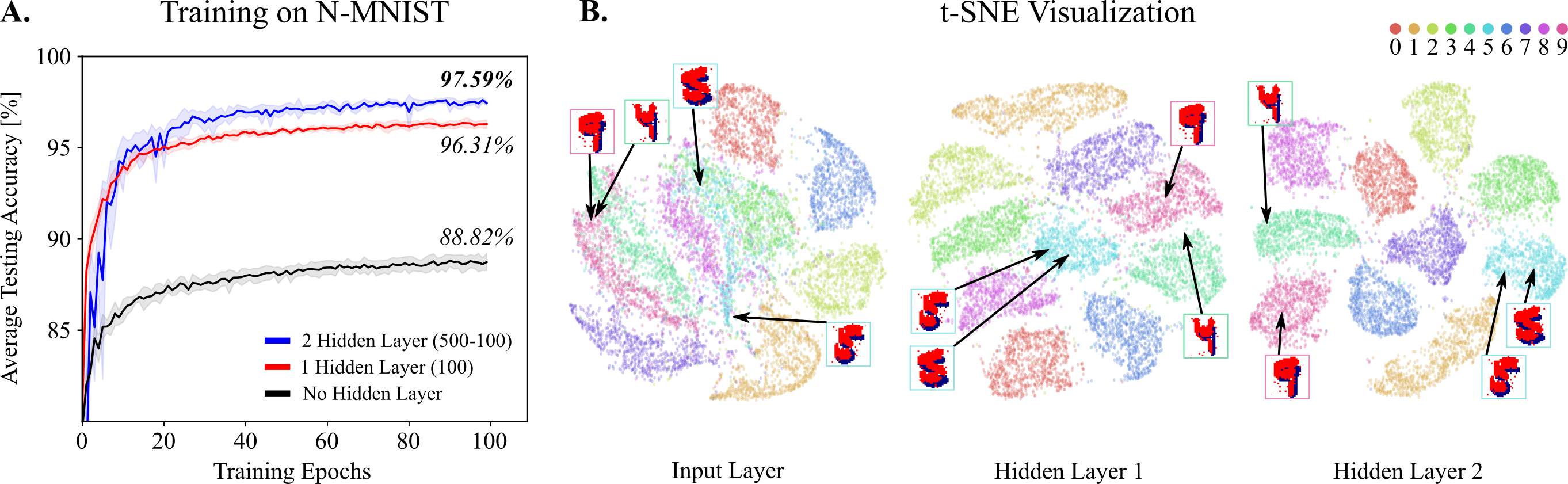}
    \caption{ Learning with multiple layers for N-MNIST. \textbf{A. }Test accuracies increased with adding more layers suggesting that deeper layers learn better representations. \textbf{B. } t-SNE reveals distinct and well-separated clusters for each digit in deeper layers. MNIST results can be found in Appendix D.}
    \label{fig:fig3}
\end{figure*}


To examine the ability of our method in taking advantage of multi-layer structure, we trained networks with zero, one, and two hidden layers. Adding more layers improved the performance of our method (Fig. \ref{fig:fig3}A), suggesting that deeper networks learned better representations of the data. We visualized the internal representations of the SNN using t-SNE~\citep{van2008visualizing} to uncover the structures in the high-dimensional manifolds. Specifically, we applied t-SNE to the spike counts of the hidden layer neurons corresponding to the input test samples in N-MNIST. Fig. \ref{fig:fig3}B, shows that deeper layers of the networks learned representations with distinct and well-separated clusters for each digit. For example, the '5' digits clustered into two distinct partitions in the input layer, formed one tight cluster in hidden layers 1 and 2. Likewise, the similarly looking digits '4' and '9' formed overlapping clusters in the input layer, which became increasingly segregated in the deeper layers.

\subsection{Effectiveness of sleep phase in SNN training}

\begin{figure*}
    \centering
    \includegraphics{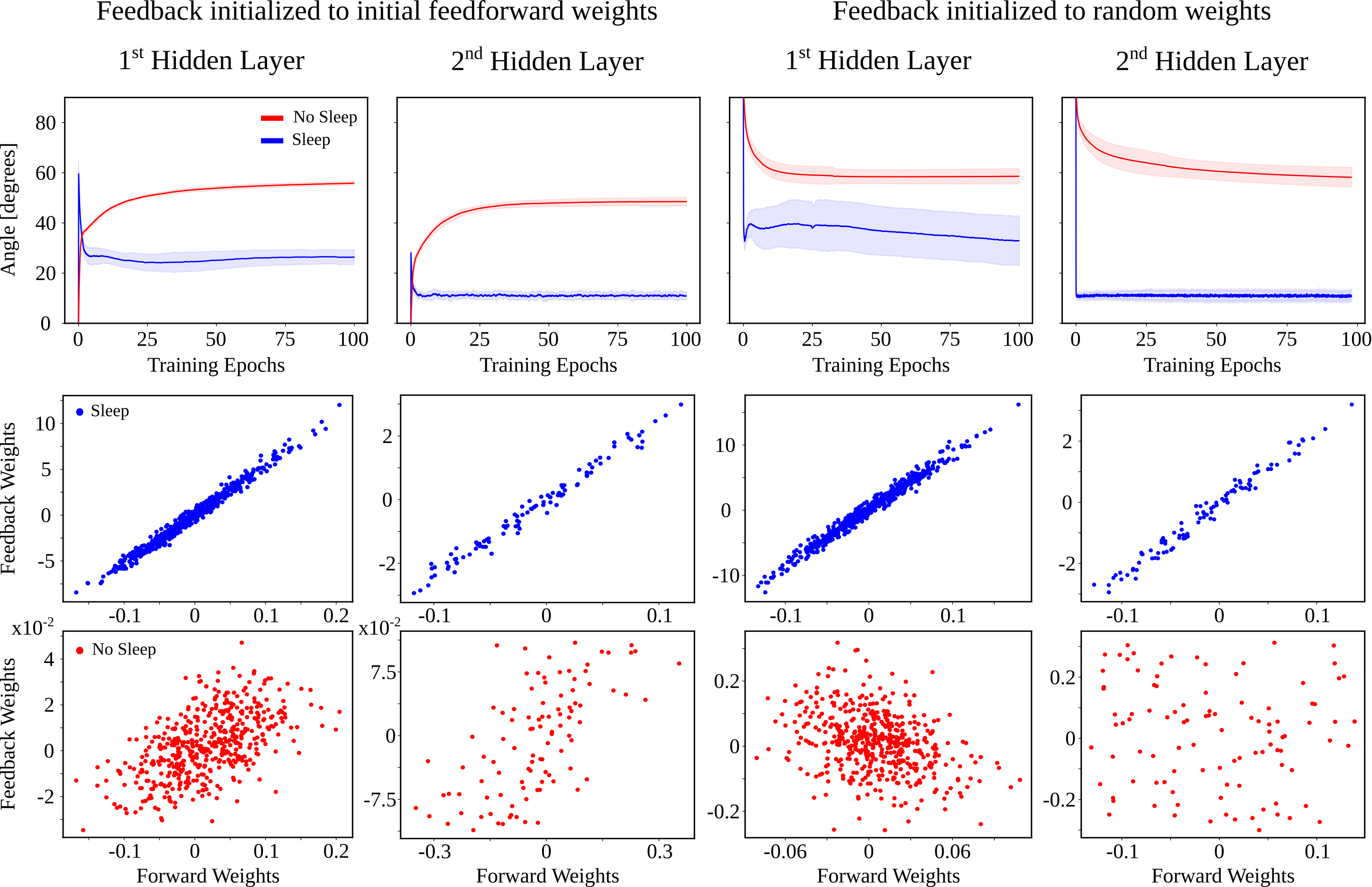}
    \caption{Effectiveness of sleep phase for N-MNIST. (Top) Networks with sleep (blue) converged to lower alignment angle between feedback and feedforward weight matrices than networks without sleep (red). (Middle) Feedback weight magnitudes for networks with sleep (blue) correlated more strongly with the magnitudes of reciprocal feedforward weights . (Bottom) This correlation was weaker for networks without sleep (red). MNIST results can be found in Appendix D.}
    \label{fig:fig4}
\end{figure*}

\begin{table*}
  \caption{Evaluating the effectiveness of sleep phase}
  \label{sleep-phase}
  \centering
  \begin{tabular}{cccc}
    \toprule
    Feedback initialized to & Feedback update & MNIST (Accuracy) & N-MNIST (Accuracy) \\
    \midrule
    Initial feedforward weights & Sleep & 98.13\%\ $\pm$\ 0.10\% & 97.59\%\ $\pm$\ 0.12\%\\
  (FwdInit) & No sleep & 97.53\%\ $\pm$\ 0.14\% & 96.10\%\ $\pm$\ 0.23\% \\
    \addlinespace[0.2cm]
    Random weights & Sleep & 98.08\%\ $\pm$\ 0.07\% & 97.42\%\ $\pm$\ 0.38\% \\
    (RandInit) & No sleep & 81.17\%\ $\pm$\ 15.83\% & 78.52\%\ $\pm$\ 14.62\% \\
    \bottomrule
  \end{tabular}
\end{table*}

To demonstrate the effectiveness of the sleep phase in training SNN, we compared our method against networks trained with fixed feedback weights. As shown in Table \ref{sleep-phase}, networks with sleep phase performed better than the ones without sleep, thereby highlighting the importance of sleep in SNN training. The improvement in performance with sleep can be attributed to the better alignment of the feedback weights with the feedforward weights. To demonstrate this, we computed the angle between the feedback and feedforward weight matrices by flattening them to vectors. As opposed to networks without sleep, networks with sleep converged to low angles for both hidden layers (Fig. \ref{fig:fig4} (top)), indicating better alignment between the feedback and feedforward weights. This can also be seen in Fig. \ref{fig:fig4} (middle and bottom), depicting the correlation between the magnitudes of feedback and reciprocal feedforward weights. When compared to networks without sleep, the feedback weight magnitudes in our method exhibited a higher correlation with the magnitude of the reciprocal feedforward weights, further demonstrating the effectiveness of sleep. 

Next, we examined if our method could utilize randomly initialized feedback weights (RandInit). Contrary to feedback weights initialized to initial feedforward weights (FwdInit), RandInit better resembles the non-identical synaptic pathways connecting upstream and downstream neurons in the brain~\citep{crick1989recent}, and offers more biological plausibility. Networks with sleep utilizing FwdInit achieved the same levels of performance as the ones utilizing RandInit (Table \ref{sleep-phase}). Consistent with our previous observation, the sleep phase improved the alignment between feedback and feedforward weights (Figure \ref{fig:fig4}). On the other hand, networks without sleep utilizing RandInit converged to higher alignment angles, suggesting that our method can indeed utilize RandInit. 

To examine how often the sleep phase was required, we trained networks with varying frequencies of sleep phase. Specifically, we put the network to sleep after every $x$ number of batches of training samples, where $x \in \{1, 16, 64, 256\}$. To ensure that the total number of sleep cycles was the same for all the cases, we set the number of sleep cycles per sleep phase equal to $x$. Table \ref{sleep-freq} shows that our method achieved the same level of performance for all sleep frequencies, suggesting that the sleep phase can be effective even if sparingly used but with a longer duration. 

\begin{table}
\small
  \caption{Evaluating the effect of sleep frequencies}
  \label{sleep-freq}
  \centering
  \begin{tabular}{cll}
    \toprule
    Training Batch & MNIST (Accuracy)     & N-MNIST (Accuracy) \\
    \midrule
    16 & 98.06\%\ $\pm$\ 0.08\% & 97.52\%\ $\pm$\ 0.11\%\\
    64 & 98.13\%\ $\pm$\ 0.05\% & 97.50\%\ $\pm$\ 0.21\%\\
    256 & 98.21\%\ $\pm$\ 0.12\% & 97.52\%\ $\pm$\ 0.24\%\\
    \bottomrule
  \end{tabular}
\end{table}


\subsection{Training on neuromorphic hardware}
To demonstrate that our algorithm can be used for on-chip learning, we deployed it on Intel's Loihi neuromorphic processor. The hardware limitations of Loihi required us to deploy a simplified version of our learning rule. The limitations and simplifications are provided in Appendix E. We trained a single hidden layer network with 100 hidden units to learn the MNIST task, and achieved $93.32\%$ test accuracy while consuming $400$ times less energy per training sample than BioGrad on GPU (detailed comparison in Appendix E). We also show the robustness of our method to non-batched and low-precision training (Appendix D), which provides further evidence that our method is generally applicable to neuromorphic hardware.

\section{Discussion}
In this work, we presented BioGrad, a biologically plausible learning algorithm for SNN that followed the neuromorphic principles of event-based, local, and online computations. Our learning rule is a close mathematical approximation of backpropagation for multi-layer networks. This approximation was further improved by the sleep phase that aligned the feedback and the feedforward weights. As a result, our method was functionally equivalent to backpropagation, whereas its robustness to non-batched and low-precision settings made it applicable to neuromorphic computing. 

The biological relevance of our learning algorithm emanates from the correspondence of several of its components to experimental neuroscientific evidence. Specifically, our multi-compartment neurons derived their structure from the pyramidal neurons in the cortex and the hippocampus~\citep{manita2015top,amaral1989three}. Similarly, our eligibility traces draw from dedicated molecules in biological neurons that capture the temporal evolution of synaptic activity and dictate synaptic plasticity~\citep{sanhueza2013camkii}. Lastly, the sleep phase improved learning by tuning the feedback weights, in alignment with the biological sleep that utilizes spontaneous neuronal activity to consolidate learning~\citep{tononi2014sleep}. 

The online nature of our learning algorithm allows it to be used in real-world applications where training data arrive in a streaming fashion, and learning needs to be performed incrementally~\citep{kading2016fine,lesort2020continual}. Further extensions may include integration with few-shot learning methods for faster learning~\citep{finn2017model,lesort2020continual}.

\bibliography{aaai22}

\clearpage
\onecolumn

\begin{center}
\textbf{\large Appendix: BioGrad: Biologically Plausible Gradient-Based Learning for Spiking Neural Networks}
\end{center}

\setcounter{section}{0}
\setcounter{figure}{0}
\setcounter{footnote}{0}
\setcounter{equation}{0}
\renewcommand{\thefigure}{S\arabic{figure}}

\section*{A. Mathematical proof of equivalence between BioGrad and backpropagation}

\subsection*{Neuron Model}

The dynamics of the somatic compartments in the SNN were governed by the leaky-integrate-and-fire (LIF) neuron model described in equation (\ref{lif:eq1}). 
\begin{equation}
\begin{gathered}
    \mathbf{v_s}^{(i)(t)} = d_v\cdot \mathbf{v_s}^{(i)(t-1)}\cdot (1-\mathbf{o_s}^{(i)(t-1)}) + \mathbf{W}^{(i)}\mathbf{o_s}^{(i-1)(t)} \\
    \mathbf{o_s}^{(i)(t)}=Threshold(\mathbf{v_s}^{(i)(t)}, V_{th})
    \label{lif:eq1}
\end{gathered}
\end{equation}

At every timestep, we aggregated the spikes of the output layer neurons as the output of the SNN:
\begin{equation}
    \mathbf{Output}^{(t)}= \mathbf{Output}^{(t-1)} + \mathbf{o_s}^{(out)(t)}
\end{equation}

\subsection*{Spatiotemporal backpropagation for SNN}

The spatiotemporal backpropagation computes the gradient of loss function $L$ with respect to weights $\mathbf{W}^{(i)}$ of layer $i$ by collecting the gradients backpropagated from all the timesteps (equation (\ref{stbp: eq1})). 
\begin{equation}
    \frac{\partial L}{\partial \mathbf{W}^{(i)}} = \sum_{t=1}^T \frac{\partial L}{\partial \mathbf{v_s}^{(i)(t)}} \frac{\partial \mathbf{v_s}^{(i)(t)}}{\partial \mathbf{W}^{(i)}}
    \label{stbp: eq1}
\end{equation}

The loss gradient can be decomposed into temporal gradients from future timesteps and spatial gradients from downstream layers using the chain rule (equation (\ref{stbp:eq2})).
\begin{equation}
    \begin{gathered}
     \frac{\partial L}{\partial \mathbf{v_s}^{(i)(t)}} = \frac{\partial L}{\partial \mathbf{v_s}^{(i)(t+1)}} \frac{\partial \mathbf{v_s}^{(i)(t+1)}}{\partial \mathbf{v_s}^{(i)(t)}} + \frac{\partial L}{\partial \mathbf{o_s}^{(i)(t)}} \frac{\partial \mathbf{o_s}^{(i)(t)}}{\partial \mathbf{v_s}^{(i)(t)}} \\
     \frac{\partial L}{\partial \mathbf{o_s}^{(i)(t)}} = \frac{\partial L}{\partial \mathbf{v_s}^{(i)(t+1)}} \frac{\partial \mathbf{v_s}^{(i)(t+1)}}{\partial \mathbf{o_s}^{(i)(t)}} + \frac{\partial L}{\partial \mathbf{v_s}^{(i+1)(t)}}\frac{\partial \mathbf{v_s}^{(i+1)(t)}}{\partial \mathbf{o_s}^{(i)(t)}}
     \label{stbp:eq2}
    \end{gathered}
\end{equation}

Combining the equations in (\ref{stbp:eq2}) and rearranging the partial derivatives to separate the temporal and spatial components results in the following form of loss gradient which is a recursive sum over time:

\begin{equation}
    \begin{aligned}
    \frac{\partial L}{\partial \mathbf{v_s}^{(i)(t)}} &= \frac{\partial L}{\partial \mathbf{v_s}^{(i)(t+1)}}(\frac{\partial \mathbf{v_s}^{(i)(t+1)}}{\partial \mathbf{v_s}^{(i)(t)}} + \frac{\partial \mathbf{v_s}^{(i)(t+1)}}{\partial \mathbf{o_s}^{(i)(t)}}\frac{\partial \mathbf{o_s}^{(i)(t)}}{\partial \mathbf{v_s}^{(i)(t)}}) + \frac{\partial L}{\partial \mathbf{v_s}^{(i+1)(t)}} \frac{\partial \mathbf{v_s}^{(i+1)(t)}}{\partial \mathbf{o_s}^{(i)(t)}}\frac{\partial \mathbf{o_s}^{(i)(t)}}{\partial \mathbf{v_s}^{(i)(t)}} \\
    &=\sum_{x=t}^T \frac{\partial L}{\partial \mathbf{v_s}^{(i+1)(x)}} \frac{\partial \mathbf{v_s}^{(i+1)(x)}}{\partial \mathbf{o_s}^{(i)(x)}}\frac{\partial \mathbf{o_s}^{(i)(x)}}{\partial \mathbf{v_s}^{(i)(x)}} \prod_{y=t}^{x-1} (\frac{\partial \mathbf{v_s}^{(i)(y+1)}}{\partial \mathbf{v_s}^{(i)(y)}} + \frac{\partial \mathbf{v_s}^{(i)(y+1)}}{\partial \mathbf{o_s}^{(i)(y)}}\frac{\partial \mathbf{o_s}^{(i)(y)}}{\partial \mathbf{v_s}^{(i)(y)}})
    \label{stbp:eq3}
     \end{aligned}
\end{equation}

where, 
\begin{equation}
    \frac{\partial \mathbf{o_s}^{(i)(t)}}{\partial \mathbf{v_s}^{(i)(t)}} \approx z(\mathbf{v_s}^{(i)(t)}) = \begin{cases}
  b & \text{if $|\mathbf{v_s}^{(i)(t)} - V_{th}| < a$} \\
  0 & \text{otherwise}
  \end{cases}
\end{equation}
where $z$ is the rectangular pseudo-gradient function to estimate the derivative of a spike~\cite{wu2018spatio}, $a$ is the psuedo-grad window and $b$ is the amplification factor.

\subsection*{Approximated form of spatiotemporal backpropagation}

To derive the synaptic learning rule of BioGrad from backpropagation, we first approximated the loss gradient of the spatiotemporal backpropagation as per equation (\ref{approx:eq1}) to alleviate the need for non-local neuronal states from downstream layers (same as equation (8) in the main text).

\begin{equation}
\frac{\partial L}{\partial \mathbf{W}^{(i)}} \approx \mathbf{B}^{(i)}\frac{\partial L}{\partial Output(T)}\sum_{t=1}^T\frac{\partial \mathbf{v_s}^{(i)(t)}}{\partial \mathbf{W}^{(i)}}\sum_{x=t}^T \frac{\partial \mathbf{o_s}^{(i)(x)}}{\partial \mathbf{v_s}^{(i)(x)}} \prod_{y=1}^{x-1} d_v\cdot \mathbf{Decay}^{(i)(y)}
\label{approx:eq1}
\end{equation}

The key approximation is that of the non-local loss gradient $\frac{\partial L}{\partial \mathbf{v_s}^{(i+1)(x)}}$ in equation (\ref{stbp:eq3}) that relies on neuronal states from downstream neurons. We approximate it by disregarding the non-local information as follows :

\begin{equation}
     \frac{\partial L}{\partial \mathbf{v_s}^{(i+1)(x)}} \frac{\partial \mathbf{v_s}^{(i+1)(x)}}{\partial \mathbf{o_s}^{(i)(x)}} \approx \prod_{j=i+1}^{K} \mathbf{W}^{(j)\top} \frac{\partial L}{\partial Output(T)} = \mathbf{B}^{(i)}\frac{\partial L}{\partial Output(T)}
\end{equation}

Further, unwrapping the inner term in the nested product of equation (\ref{stbp:eq3}) gives us: 
\begin{equation}
    \begin{gathered}
     \frac{\partial \mathbf{v_s}^{(i)(z+1)}}{\partial \mathbf{v_s}^{(i)(z)}} + \frac{\partial \mathbf{v_s}^{(i)(z+1)}}{\partial \mathbf{o_s}^{(i)(z)}}\frac{\partial \mathbf{o_s}^{(i)(z)}}{\partial \mathbf{v_s}^{(i)(z)}} = d_v \cdot \mathbf{Decay}^{(i)(z)} \\
     \mathbf{Decay}^{(i)(t)} = 1-\mathbf{o_s}^{(i)(t-1)}-\mathbf{v_s}^{(i)(t-1)}\cdot z(\mathbf{v_s}^{(i)(t-1)})
     \label{approx:eq2}
    \end{gathered}
\end{equation}
It should be noted that $\mathbf{Decay}^{(i)(t)}$ relies on only local neuronal states from past timesteps. 

\subsection*{Deriving BioGrad from approximated spatiotemporal backpropagation}

We derive BioGrad from the approximated spatiotemporal backpropagation by evaluating and rearranging the partial derivatives in equation (\ref{approx:eq1}) so that the inner sum operates only over past timesteps (not relying on future information) as per equation (\ref{biograd:eq1}). This step is similar to the method proposed in~\cite{bellec2020solution}.

\begin{equation}
    \begin{aligned}
    \frac{\partial L}{\partial \mathbf{W}^{(i)}} &\approx \mathbf{B}^{(i)}\frac{\partial L}{\partial Output(T)}\sum_{t=1}^T\mathbf{o_s}^{(i-1)(t)}\sum_{x=t}^T z(\mathbf{v_s}^{(i)(x)}) \prod_{y=1}^{x-1} d_v\cdot \mathbf{Decay}^{(i)(y)} \\
    &= \mathbf{B}^{(i)}\frac{\partial L}{\partial Output(T)}\sum_{t=1}^T z(\mathbf{v_s}^{(i)(t)}) \sum_{x=1}^t \mathbf{o_s}^{(i-1)(x)} \prod_{y=x}^{t-1} d_v\cdot \mathbf{Decay}^{(i)(y)}
    \label{biograd:eq1}
    \end{aligned}
\end{equation}

Next, we define the presynaptic spike trace as:
\begin{equation}
    \begin{aligned}
    \mathbf{Tr_{pre}}^{(i)(t)} &= \sum_{x=1}^t \mathbf{o_s}^{(i-1)(t)} \prod_{y=x}^{t-1} d_v\cdot \mathbf{Decay}^{(i)(y)} \\
    &= d_v\cdot \mathbf{Decay}^{(i)(t)} \cdot \mathbf{Tr_{pre}}^{(i)(t-1)} + \mathbf{o_s}^{(i-1)(t)}
    \end{aligned}
\end{equation}

and the correlation trace as:
\begin{equation}
    \begin{aligned}
    \mathbf{Tr_{corr}}^{(i)(r)} &= \sum_{t=1}^r z(\mathbf{v_s}^{(i)(t-1)}) \mathbf{Tr_{pre}}^{(i)(t)} \\
    &= \mathbf{Tr_{corr}}^{(i)(r-1)} + \mathbf{Tr_{pre}}^{(i)(r)}\cdot z(\mathbf{v_s}^{(i)(r-1)})
    \end{aligned}
\end{equation}

The spatial gradient corresponds to the membrane voltage of the apical compartment:
\begin{equation}
    \frac{\mathbf{v_a}^{(i)(T)}}{T-t_{error}} \approx \mathbf{B}^{(i)}\frac{\partial L}{\partial Output(T)}
\end{equation}

Finally, the weights are updated by the synaptic learning rule of BioGrad as:
\begin{equation}
    \Delta \mathbf{W}^{(i)} = -\eta\cdot \frac{\mathbf{v_a}^{(i)(T)}}{T-t_{error}}\cdot \mathbf{Tr_{corr}}^{(i)(T)}
\end{equation}
\section*{B. Zero-mean random inputs in sleep phase}
We show here how the use of zero-mean random inputs in the sleep phase can allow the feedback weight update to grow in the direction of feedforward weights ~\cite{akrout2019deep}:

Let $x$ be the random input to the layer for which we are updating the feedback weight, and let $y$ be the corresponding network output. With an assumption of random zero-mean input, we can write:
\begin{equation}
    y = W^T \cdot x
\end{equation}

\begin{equation}  
E[xy^T] = E[x.x^T.W^T] = \sigma^2 W^T
\end{equation}

With the feedback weight update equation being:
\begin{equation}
\Delta B = \eta * x * y^T
\end{equation}

From equations 16 and 17, we can see that with the assumptions that we have, the feedback weight B grows in the direction of the feedforward weights. 

In our implementation, we achieved the zero-mean condition by utilizing random positive and negative spikes. Interestingly, the distinct neuronal representations of positive and negative values are biologically plausible, with negative value-coding neurons having been identified in several brain areas, including the ocular system \cite{leigh2015neurology} and are associated with efficient learning in the amygdala \cite{paton2006primate}.

\section*{C. List of hyperparameters}
Here, we describe the hyperparameter configurations for training SNN with BioGrad and the baseline backprop methods. The baselines used the same hyperparameters as the BioGrad unless explicitly stated. Hyperparameter configurations for all methods were as follows:
\begin{itemize}
    \item \textbf{Neuron Parameters}
    \begin{itemize}
        \item Voltage decay: $0.6$ (MNIST); $0.3$ (N-MNIST) 
        \item Voltage threshold: $0.3$
        \item Pseudo-gradient function parameters: window $0.3$; amplification factor $1.0$
    \end{itemize}
    \item \textbf{Training SNN with BioGrad (MNIST)}
    \begin{itemize}
        \item[$-$] Network architecture: 3-layered $[(28*28)-500-100-10]$
        \item[$-$] Feedforward weight learning rate: $1e-3$ 
        \item[$-$] Sleep phase learning rate: $(1e-4)/3$
        \item[$-$] Sleep frequency: 1 sleep cycle after every training batch
        \item[$-$] Sample presentation timesteps: $20$
        \item[$-$] Random input presentation timesteps during sleep: $50$
        \item[$-$] Batch size: $128$ (Train); $128$ (sleep)
        \item[$-$] Starting timestep for feeding spike-encoded errors to the error neurons: $t_{error}=5$
    \end{itemize}
    
    \item \textbf{Training SNN with BioGrad (N-MNIST)}
    \begin{itemize}
        \item[$-$] Network architectures: 3-layered $[(34*34*2)-500-100-10]$
        \item[$-$] Feedforward weight learning rate: $1e-3$  
        \item[$-$] Sleep phase learning rate: $(1e-4)/3$
        \item[$-$] Sleep frequency: 1 sleep cycle after every training batch
        \item[$-$] Sample presentation timesteps: $60$
        \item[$-$] Random input presentation timesteps during sleep: $50$
        \item[$-$] Batch size: $128$ (Train); $128$ (sleep)
        \item[$-$] Starting timestep for feeding spike-encoded errors to the error neurons: $t_{error}=19$
    \end{itemize}
    
    \item \textbf{Training SNN with Backprop (MNIST)}
    \begin{itemize}
        \item[$-$] Feedforward weight learning rate: $5e-4$ (Adam); $0.9e-2$ (SGD)
    \end{itemize}
    
    \item \textbf{Training SNN with Backprop (N-MNIST)}
    \begin{itemize}
        \item[$-$] Feedforward weight learning rate: $1e-4$ (Adam); $0.9e-2$ (SGD)
    \end{itemize}
\end{itemize}

\section*{D. Additional results}

\subsection*{Additional MNIST results}
\begin{figure}[!h]
    \centering
    \includegraphics[scale=1.5]{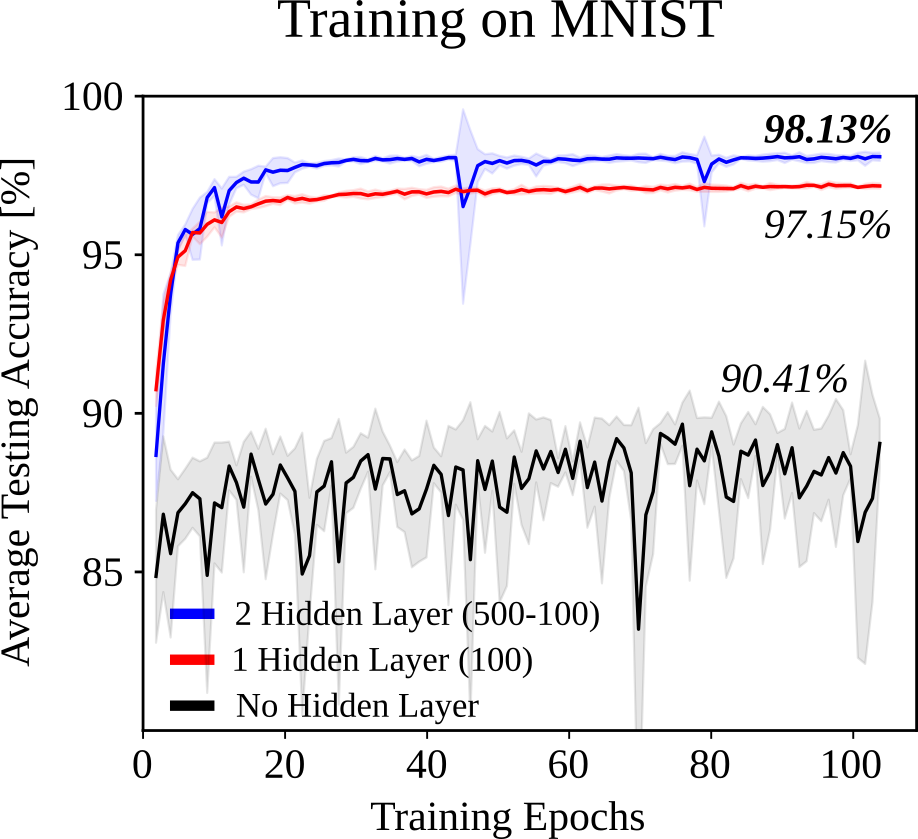}
    \caption{Learning with multiple layers for MNIST. Test accuracies increased with adding more layers suggesting that deeper layers learn better representations.}
    \label{fig:sfig3}
\end{figure}

\begin{figure}[!h]
    \centering
    \includegraphics{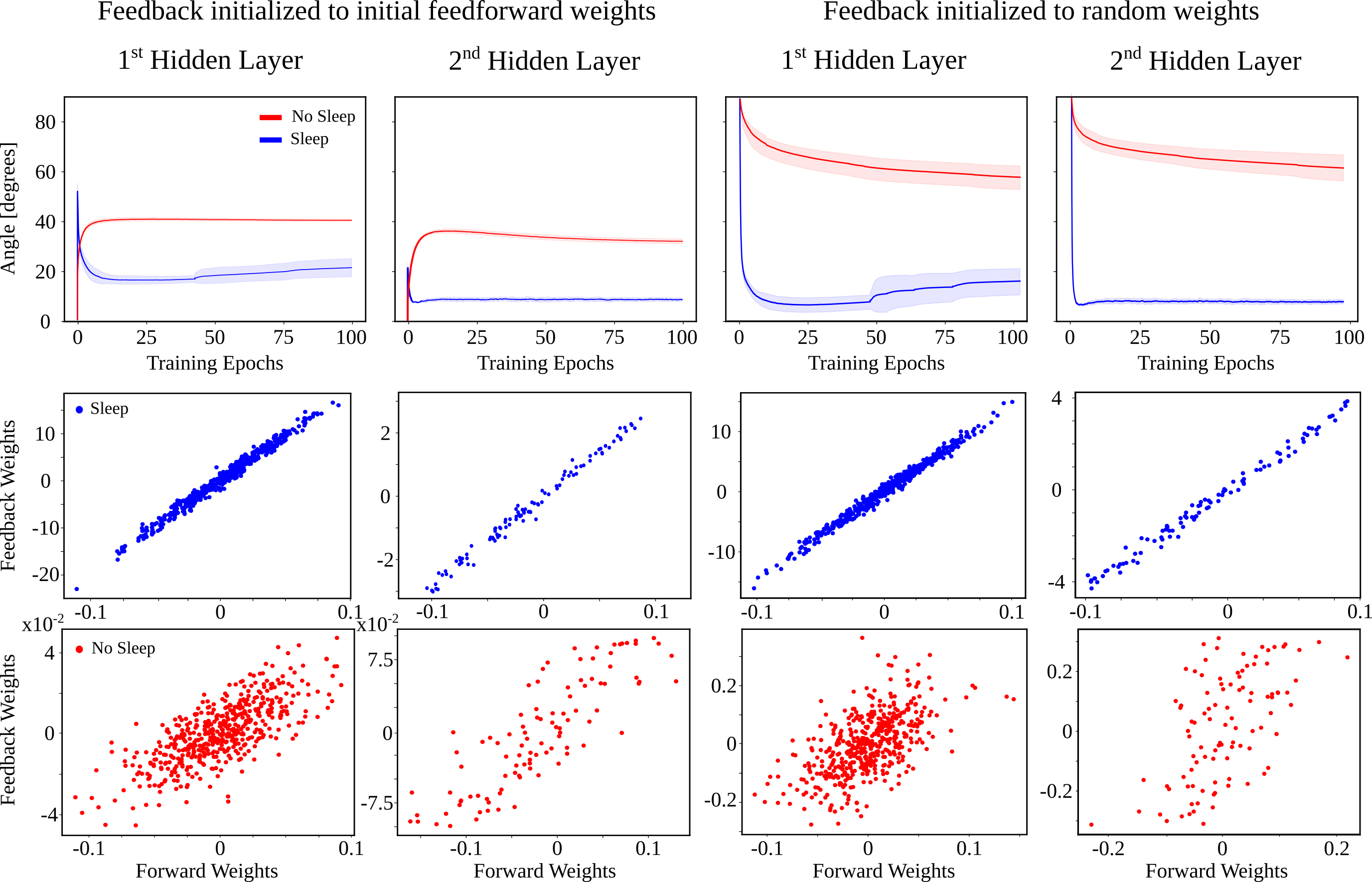}
    \caption{Effectiveness of sleep phase for MNIST. (Top) Networks with sleep (blue) converged to lower alignment angle between feedback and feedforward weight matrices than networks without sleep (red). (Middle) Feedback weight magnitudes for networks with sleep (blue) correlated more strongly with the magnitudes of reciprocal feedforward weights . (Bottom) This correlation was weaker for networks without sleep (red).}
    \label{fig:sfig4}
\end{figure}

We show the MNIST results for multi-layer ablation in Fig. \ref{fig:sfig3} and sleep phase effectiveness in Fig. \ref{fig:sfig4}. 

\subsection*{Training in non-batched setting}
To examine if our method can work in a non-batched setting, we performed weight updates after each training sample. The non-batched training achieved the same level of performance as the mini-batch training for both MNIST ($98.23\%\ \pm\ 0.09\%$) and N-MNIST($97.57\%\ \pm\ 0.10\%$), suggesting that our method is robust to the non-batched setting. This makes it suitable for neuromorphic computing and computing on the edge, where training samples arrive one by one and not in batches. 

\subsection*{Training Low-precision SNN and Validation on a Neuromorphic Chip}
We trained networks with quantized synaptic weights and traces to simulate the low-precision characteristic of neuromorphic hardware. For quantization, we rounded the synaptic weights and traces to the target precision using the deterministic rounding method~\cite{kahan1996ieee} whenever the variables changed their values. Even when the precision was reduced to as low as 8 bits, our method achieved a high level of performance (Table \ref{low-precision}). The low-precision of the trained model allowed us to deploy it on Intel's Loihi neuromorphic chip for inference, achieving $96.38\%\ \pm\ 0.34\%$  accuracy on MNIST.

\begin{table*}
  \caption{Evaluating the effect of low-precision learning}
  \label{low-precision}
  \centering
  \begin{tabular}{cll}
    \toprule
    Precision    & MNIST (Accuracy)     & N-MNIST (Accuracy) \\
    \midrule
    32-bits & 98.13\%\ $\pm$\ 0.10\% & 97.59\%\ $\pm$\ 0.12\%\\
    16-bits & 98.10\%\ $\pm$\ 0.07\% & 97.50\%\ $\pm$\ 0.08\%\\
    8-bits & 96.72\%\ $\pm$\ 0.20\% & 95.34\%\ $\pm$\ 0.26\% \\
    \bottomrule
  \end{tabular}
\end{table*}

\section*{E. Training on Neuromorphic Hardware}
The hardware limitations of Loihi required us to deploy a simplified version of our learning rule. Specifically, Loihi does not support learning within a multi-compartment neuron and the learning rule can only use spikes but not other neuronal states. Due to this, we used the spiking activity of the neurons as our pseudograd function, and turned $\mathbf{v_a}$ into a spiking unit. Since we could not deploy the sleep phase due to the limitation, we set the feedback weights to feedforward weights after every epoch of training to simulate the sleep process. Moreover, since the traces on loihi have very low precision ($< 4 bits$), we used the Integrate-and-Fire (IF) model for the neuron instead of LIF to make the learning robust to less accurate estimation of traces.

We measured the average power consumed and the speed of training for the MNIST task (Table \ref{loihi-power}). For power measurement, we used software tools that probed the on-board sensors of each device: nvidia-smi for GPU (RTX 3090) and energy probe for Loihi (Nahuku32). The energy consumption per training image was computed by multipling the dynamic power consumption and the training speed.

\begin{table*}
  \caption{Power performance and training speed across hardware}
  \label{loihi-power}
  \centering
  \begin{tabular}{ccccc}
    \toprule
    Device & Static (W) & Dynamic (W) & Speed (ms/image) & Energy(mJ/image)\\
    \midrule
    GPU & 32 & 94 & 4.54 & 426.76\\
    Loihi & 1.035 & 0.056 & 18.83 & 1.054\\
    \bottomrule
  \end{tabular}
\end{table*}

\section*{F. Hardware details}
For training the networks, we employed internal clusters of Nvidia Tesla K40 GPUs with 12Gb RAM and a Nvidia RTX 3090 GPU with 24Gb RAM.

\end{document}